\documentclass[11pt]{article}
\usepackage{enumitem}
\usepackage{algorithm}
\usepackage{algorithmic}
\usepackage{amsmath}
\usepackage{amsfonts}
\usepackage{dsfont}
\usepackage{booktabs}
\usepackage{multirow}
\usepackage{makecell}
\usepackage{tabularx}
\usepackage{siunitx} 

\usepackage{graphicx} 

\usepackage[preprint]{acl}

\usepackage{times}
\usepackage{latexsym}

\usepackage[T1]{fontenc}

\usepackage[utf8]{inputenc}

\usepackage{microtype}

\usepackage{inconsolata}

\title{TETHER: Training-Free, Elastic, ToC-Anchored Hierarchy Extraction and Recovery in Design-Driven Long Documents}

\author{
  \textbf{Yu Zou\textsuperscript{1,*}},
  \textbf{Yan Chen\textsuperscript{1,*}},
  \textbf{Jialei Zeng\textsuperscript{1}},
  \textbf{Xiaorui Zhou\textsuperscript{1}},
  \textbf{Aixi Zhong\textsuperscript{1}}
  \\
  \textbf{Qi Zhou\textsuperscript{1}},
  \textbf{Haoran You\textsuperscript{1}},
  \textbf{Ke Jiang\textsuperscript{2}},
  \textbf{Changyu Hu\textsuperscript{3,\textdagger}}
  \\[0.4em]
  {\normalfont
  \textsuperscript{1}Southwestern University of Finance and Economics,
  Chengdu, Sichuan, China}
  \\
  {\normalfont
  \textsuperscript{2}Pujiang County Bureau of Transportation,
  Chengdu, Sichuan, China}
  \\
  {\normalfont
  \textsuperscript{3}
  Ningbo University of Finance and Economics,
  Ningbo, Zhejiang, China}
  \\[0.2em]
  {\normalfont\small
  \textsuperscript{*}Equal contribution.
  \quad
  \textsuperscript{\textdagger}Corresponding author:
  \href{mailto:huchangyu@nbufe.edu.cn}
       {\texttt{huchangyu@nbufe.edu.cn}}.}
}

\begin{document}
\maketitle
\begin{abstract}
Converting complex documents into structured knowledge constitutes the critical infrastructure urgently required for deploying Large Language Models (LLMs) in downstream research.
However, blind to the Table of Contents (ToC) as a macro-structure, existing bottom-up paradigms rely on localized visual cues, inducing heuristic fragility in high-entropy layouts where visual prominence decouples from logical depth; meanwhile, direct LLM extraction lacks spatial grounding, triggering hallucinations in long contexts.
To address this, we introduce \textbf{TETHER}\footnote{The source code is publicly available on
\href{https://github.com/liucun-zy/Hierarchical-Structure-Restoration-for-Complex-Long-Documents}{GitHub}.}, a training-free framework that elevates the ToC from discarded noise to a vital structural prior, enabling robust adaptation across design-driven long documents.
TETHER operates via a two-stage neuro-symbolic mechanism: first, it constructs a document skeleton from the raw ToC using a Large Multimodal Model (LMM) empowered by spatio-semantic constraints; second, it employs an anchor-based alignment strategy to tether content blocks to this skeleton, effectively resolving boundary ambiguities.
Evaluated across a multilingual benchmark of ESG reports, product manuals, and annual reports, TETHER delivers highly precise, state-of-the-art (SOTA) zero-shot hierarchy recovery, outperforming deep hierarchical recovery models, dedicated parsers, and general-purpose LMMs.
Moreover, stress tests highlight TETHER's resilience to sparse omissions and textual noise, demonstrating that its precise semantic alignment is fundamentally anchored in topological and depth fidelity.
Crucially, ablations confirm the framework's architectural irreducibility, while achieving ~90\% token reduction and drastic speedups over end-to-end baselines, establishing a compelling equilibrium between structural precision and deployment scalability.
\end{abstract}

\section{Introduction}

Deploying Large Language Models (LLMs) and Retrieval-Augmented Generation (RAG) in real-world domains heavily relies on processing massive, complex documents. To unlock their reasoning potential, 
converting these unstructured layouts spanning hundreds of pages into rigorous hierarchical formats has become an urgent prerequisite, 
providing the essential infrastructure to navigate deep cross-page dependencies.

However, current paradigms struggle to maintain structural coherence in long-form documents. Bottom-up models (e.g., deep hierarchical recovery models~\cite{hu2022multimodal, wang2023scalable} and dedicated parsers~\cite{niu2026mineru2, marker2025}) rely on local visual cues to infer structure, leading to severe structural distortion whenever visual formatting contradicts logical hierarchy. Moreover, such specialized approaches are inherently prone to domain overfitting. Concurrently, direct LLM extraction lacks spatial grounding, inducing hallucinations across unanchored extended contexts~\cite{perot2024lmdx, liu2025towards}.

To mitigate the challenges of extended lengths and high-entropy layouts in documents, a deterministic global map is imperative. Fortunately, virtually all ultra-long documents conventionally feature a native structural blueprint: the ToC. Surprisingly, this structural prior remains largely unexploited in prior work, as real-world ToCs are frequently structurally implicit, typographically heterogeneous, and severely corrupted by noise.

Therefore, we introduce \textbf{TETHER} (\textbf{T}raining-Free, \textbf{E}lastic, \textbf{T}oC-Guided \textbf{H}ierarchy \textbf{E}xtraction and \allowbreak \textbf{R}ecovery), an architecture tailored for long-form documents with complex hierarchical structures. Rather than treating the ToC as extraneous information~\cite{hu2022multimodal,wang2023scalable}, TETHER formulates it as a native structural prior, serving as an explicit reference schema to anchor cross-page content alignment.
TETHER operates through a two-stage neuro-symbolic mechanism. In the first stage, a Large Multimodal Model (LMM) guided by spatio-semantic constraints extracts a rigorous document skeleton from the raw ToC. In the second stage, an anchor-based alignment strategy tethers content blocks to this skeleton to resolve boundary ambiguities.

To evaluate TETHER's effectiveness, generalizability, and architectural necessity, we curate a multilingual benchmark (English, Simplified, and Traditional Chinese) comprising annual reports (proxying rigidly structured, compliance-driven documents), ESG reports (representing layout-flexible \& unstructured publications), and product brochures (mirroring fragmented technical literature). This heterogeneity (Appendix~\ref{sec:challenges}) benchmarks zero-shot structural recovery across unseen layouts.

Empirically, TETHER outperforms deep hierarchical recovery models (MTD~\cite{hu2022multimodal} and CMM~\cite{wang2023scalable}), dedicated parsers (MinerU2-VLM~\cite{niu2026mineru2}, Marker~\cite{marker2025}, Docling~\cite{livathinos2025docling}, and TextIn~\cite{textin_xparse_docs}), and general-purpose LMMs (Qwen3.5-397B-A17B, GPT-5.5, Gemini3.1-Pro, and DeepSeekOCR). 
Ablations confirm its architectural necessity, while efficiency analyses reveal ~90\% token and ~80\% runtime reductions versus end-to-end baselines, striking a balance of precision and scalability.
Moreover, stress tests highlight TETHER's resilience to sparse omissions and textual noise, demonstrating that its precise semantic alignment is fundamentally anchored in topological and depth fidelity.

Our main contributions are as follows:

\textit{\textbf{(1) Paradigm Shift.}} Reversing the reliance on bottom-up local aggregation, we propose TETHER to reformulate document parsing as ToC-anchored structural grounding. Driven by top-down constraint satisfaction, this paradigm strictly confines outputs within the document's native hierarchy, neutralizing the risks of unconstrained extraction.

\textit{\textbf{(2) Methodological Novelty.} }
Central to our approach is a training-free neuro-symbolic architecture with adaptive anchoring. Synergizing LMM-based ToC extraction and iterative content alignment, it ensures topological consistency, conquers spatial blindness, and achieves zero-shot universality across diverse domains.

\textit{\textbf{(3) Empirical Superiority.}} Evaluated across three diverse categories of highly irregular multilingual documents, TETHER demonstrates SOTA superiority, resilience to intermediate skeleton perturbations, and architectural irreducibility, while achieving drastic token and runtime reductions against end-to-end baselines.

\section{Related Work}
Recent work increasingly treats long documents not as flat OCR streams, but as structured objects whose logical hierarchy determines whether downstream LLM and RAG systems can navigate them reliably~\cite{ma2024mmlongbench, livathinos2025docling, jeong2025zero}. Existing structure recovery methods broadly fall into two paradigms.

\paragraph{\textbf{Content-Driven Hierarchy Prediction.}}
Bypassing ToC priors, these methods extract structural hierarchies directly from page content. First, specialized heuristic parsers (e.g., Docling~\cite{livathinos2025docling}, MinerU~\cite{wang2024mineru}, Marker~\cite{marker2025}, TextIn~\cite{textin_xparse_docs}) couple layout analysis with OCR engines~\cite{easyocr_repo, cui2025paddleocr, smith2007overview}, deriving logical hierarchy by applying predefined rules to visual typographic cues (e.g., font size, indentation). Second, deep hierarchical models (e.g., MTD~\cite{hu2022multimodal}, CMM~\cite{wang2023scalable}) train neural networks to infer hierarchical sequences from learned typographic and layout priors. Third, training-free generative approaches such as DocsRay~\cite{jeong2025zero} bypass explicit layout analysis entirely, prompting large multimodal models (LMMs) to synthesize a pseudo-ToC solely from raw page content.
Crucially, under high-entropy layout conditions, visual prominence frequently decouples from logical structural roles, leading to systematic failures in both heuristic and neural parsers~\cite{pfitzmann2022doclaynet, da2023vision}. Notably, CMM reports that over 10.8\% of headers deviate from learned typographic patterns, indicating typographic overfitting and limited out-of-distribution generalization~\cite{wang2023scalable}. Meanwhile, lacking spatial grounding, LMM-based generative methods inevitably hallucinate hierarchical boundaries when processing extended contexts~\cite{jeong2025zero}. 

\begin{figure*}[ht!]
  \centering
  \includegraphics[width=\textwidth]{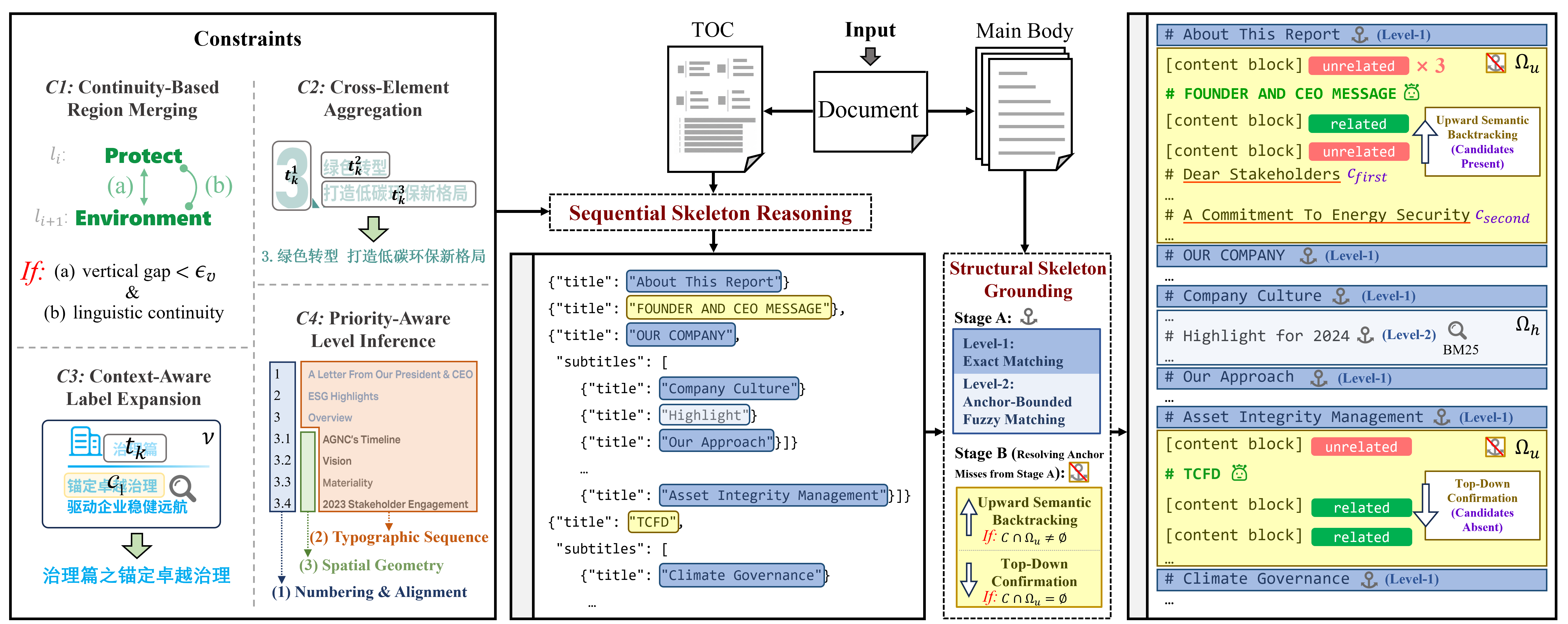}
 
  \caption{Overview of the TETHER. (1) \textit{Sequential Skeleton Reasoning} distills ToC structural priors under spatio-semantic constraints (C1--C4). (2) \textit{Structural Skeleton Grounding} aligns this skeleton to the body via a coarse-to-fine neuro-symbolic pipeline of symbolic matching (Stage A) and agentic inference (Stage B).}
  \label{fig:overview}
\end{figure*}

\paragraph{\textbf{Metadata-Driven Extraction.}}
This paradigm avoids noisy content-level prediction by recovering hierarchy from pre-existing digital signals, such as PDF tags, outlines, or XML-derived heading features~\citep{iso2017document,adobe2026bookmarks}. 
For example, HiPS segments legal textbooks by matching PDF outline metadata to the full text~\cite{wehnert2025hips}. However, this assumes noise-free metadata and fails to handle discrepancies between the metadata and the main text. 
Notably, constrained by extreme layout heterogeneity, parsing the visual ToC as a structural prior to guide body alignment has been a widely avoided paradigm~\cite{wang2023scalable, jeong2025zero, wehnert2025hips}.
TETHER pioneers this approach, addressing two core challenges: (1) Hierarchy extraction against severe ToC layout entropy, and (2) Hierarchy recovery to resolve complex ToC--body textual discrepancies.

\section{Methodology}

TETHER's pipeline (Figure~\ref{fig:overview}) comprises \textit{Sequential Skeleton Reasoning} to repurpose the ToC as a deterministic structural prior, and \textit{Structural Skeleton Grounding} for precise content alignment.

\subsection{Sequential Skeleton Reasoning}

We formulate the extraction of the target skeleton $\mathcal{H}$ from raw ToCs as an LMM-driven sequential reasoning task. This process is governed by universal structural priors, formally defined as an orthogonal framework of four constraints that systematically decouple layout anomalies across spatial (C1, C2), semantic (C3), and hierarchical (C4) dimensions:

\textit{\textbf{C1}: Continuity-Based Region Merging.} To resolve header fragmentation in narrow-column layouts, adjacent lines $l_i$ and $l_{i+1}$ are merged if they satisfy spatial proximity (vertical gap $< \epsilon_v$, scaled by dominant line height) and linguistic continuity constraints, yielding unified semantic units.

\textit{\textbf{C2}: Cross-Element Aggregation.} To mitigate visual scattering in complex layouts, distributed title segments (e.g., numbers, core texts, and tags) are unified into a composite title $t_k$ based on horizontal alignment and layout grouping signals (e.g., shared backgrounds).

\textit{\textbf{C3}: Context-Aware Label Expansion.} To resolve semantic sparsity in labels $t_k$ (e.g., ``Part A''), descriptive candidates are harvested from a proximal visual window $\mathcal{V}$. Candidates are validated via composite Jaccard and normalized Levenshtein metrics to synthesize a semantically complete title.

\textit{\textbf{C4}: Priority-Driven Level Inference.} Subordinating unstable visual cues (e.g., the ``larger-is-higher'' rule) to structural logic, we infer hierarchical level $\ell_k$ via a priority-ordered cascade: (1) \textit{Numbering \& Alignment} to validate nesting and sibling relationships; (2) \textit{Typographic Sequence} incorporating weight, size, and color to resolve ambiguous ties; and (3) \textit{Spatial Geometry} utilizing indentation and gaps to delineate parent boundaries.

\paragraph{\textbf{Skeleton Formalization.}}

Ultimately, this reasoning chain yields a purified hierarchical sequence:
{
\setlength{\abovedisplayskip}{4pt} 
\setlength{\belowdisplayskip}{4pt} 
\begin{equation}
\mathcal{H} = ( h_k = (t_k, \ell_k) )_{k=1}^K
\end{equation}
}where each entry $h_k$ couples the reconstructed text $t_k$ (via C1--C3) with its inferred depth $\ell_k \in \mathbb{N}$ (via C4). Crucially, the index $k$ strictly follows the document's natural reading order thanks to the LMM's sequential generation. This ordered skeleton $\mathcal{H}$ then serves as the deterministic anchor for the downstream alignment phase.

\subsection{Structural Skeleton Grounding}
To ground the structured skeleton $\mathcal{H}$ onto the linear document stream $D$, we propose a two-stage neuro-symbolic protocol: (1) \textit{Symbolic Deterministic Anchoring} establishes a definitive structural backbone, followed by (2) \textit{Neural Agentic Fine Insertion} for context-aware boundary calibration.

\subsubsection{Task Formulation: Topology-Constrained Insertion}
Formally, given the document stream \(D = (b_n)_{n=1}^N\) and visual candidates \(\mathcal{C} \subset D\), we formulate alignment as a latent structure insertion task. Conventional extraction methods only work with observed candidates, while we cast alignment into a topology-constrained insertion process. Each node \(h_k \in \mathcal{H}\) is anchored within a restricted search space, where parent-child hierarchy and sequential order serve as hard constraints. Benefiting from this formulation, our method infers latent boundaries within valid topological windows to locate ghost nodes, leveraging structural reasoning to compensate for insufficient visual signals.

To avoid large-scale annotation, we propose a training-free framework that models this insertion as a constraint satisfaction problem (CSP) anchored by $\mathcal{H}$. This formulation fundamentally decouples topological integrity from semantic reasoning: symbolic logic provides mathematical guardrails, ensuring the neural agent operates exclusively within a provably safe sub-space. Consequently, this decoupled architecture yields three system-level guarantees:

\begin{itemize}[itemsep=0pt, parsep=0pt, topsep=0pt]
    \item \textit{Zero-Shot Generalization:} Symbolic logic and few-shot reasoning enable direct domain adaptation without fine-tuning.
    \item \textit{Bounded Inference:} Anchor-defined local windows reduce search complexity from linear scanning to localized queries.
    \item \textit{Topology-Preserving Integrity:} Global hierarchy constraints strictly prevent logical violations during boundary inference.
\end{itemize}

\subsubsection{Stage A: Symbolic Anchoring}
This stage establishes a definitive structural backbone via a precision-first, two-level symbolic protocol to anchor high-confidence matches in $\mathcal{C}$.

\paragraph{\textbf{Level-1:} Global Exact Matching.}
We initialize the anchor set $\mathcal{M}$ with normalized exact matches ($\text{Norm}(c_j) = \text{Norm}(h_k)$) from $\mathcal{C}$ for each $h_k \in \mathcal{H}$.

\paragraph{\textbf{Level-2:} Anchor-Bounded Fuzzy Matching.}
For each unmatched $h_k$, we define a window $\Omega_h$ bounded by its nearest anchors in $\mathcal{M}$. Within $\Omega_h$, we apply BM25 retrieval over $\mathcal{C}$ to resolve noisy variations (e.g., ``Env.'' to ``Environmental''). These matches augment $\mathcal{M}$, leaving the residual \textit{ghost nodes} $\mathcal{U}$ for the neural phase.

Crucially, this stage reduces intractable global search to $\Omega$-bounded inference. This topological firewall mathematically restricts the subsequent neural reasoning space, preventing local errors from cascading into structural hallucinations.

\subsubsection{Stage B: Neural Agentic Fine Insertion}
For each ghost node $u \in \mathcal{U}$, the neural agent pinpoints its optimal semantic onset $p^*$ within an $\mathcal{M}$-anchored constraint window $\Omega_u$ in two steps:

\paragraph{\textbf{Step 1:} Topology-Constrained Initialization.}
We establish a tentative anchor $p_{init}$ for $u$, adapting to the presence of visual candidates ($\mathcal{C} \cap \Omega_u$):

\noindent \textit{\textbf{Scenario I ($\mathcal{C} \cap \Omega_u \neq \emptyset$):}} 
The agent scans the candidate subset for semantic matches, anchoring $p_{init}$ to the upper boundary of the first match $c_{first}$.

\noindent \textit{\textbf{Scenario II ($\mathcal{C} \cap \Omega_u = \emptyset$):}} 
Lacking candidates, $p_{init}$ deterministically defaults to immediately below the window's upper bound $\Omega_{start}$.

\paragraph{\textbf{Step 2:} Semantic Boundary Calibration.}
From \(p_{init}\), the agent eschews unconstrained generation for \(\Omega\)-gated, state-aware traversals to pinpoint the semantic onset \(p^*\) via scenario-specific strategies:

\textit{\textbf{Calibration for Scenario I (Upward Semantic Backtracking):}}
To capture introductory content strictly preceding the candidate heading, the agent iteratively scans text blocks upward from $p_{init}$. Upon detecting semantic relevance to $u$, $p^*$ is re-anchored to the current block's upper boundary. Backtracking terminates upon encountering three consecutive unrelated blocks from the previous section, finalizing $p^*$ at their lower boundary.

\textit{\textbf{Calibration for Scenario II (Top-Down Confirmation):}}
The agent executes a semantic scan downstream from $p_{init}$, bypassing transitional noise or irrelevant segments belonging to the previous section. The process terminates upon validating two consecutive blocks relevant to $u$, positioning $p^*$ at the upper boundary of the first valid block.

Finally, recovered nodes anchored at $p^*$ augment $\mathcal{M}$ to complete the mapping.
Crucially, bounded by $\Omega$, the agent transcends exact string matching to achieve robust semantic alignment. By analyzing contextual semantics, it successfully resolves complex paraphrasing (e.g., aligning the body's ``Leader's Speech'' with the ToC's ``Chairman's Address'') without risking the downstream structural hallucinations typically triggered by unbounded LLM extraction.

\begin{table*}[t]
\centering
\caption{Hierarchy extraction performance (\%). Main values reflect our proposed \textit{\textbf{constrained extraction}} (an orthogonal 4-constraint framework); bracketed values (\textcolor{teal}{+$\Delta$}) indicate gains over prior-free \textit{\textbf{naive extraction}}. \textit{\textbf{Ours}} applies TETHER on the compact Qwen2.5-VL, enabling secure, local deployment on limited resources.}

\label{tab:toc_extraction_results}

\small 
\setlength{\tabcolsep}{3pt} 
\renewcommand{\arraystretch}{1.2} 

\newcommand{\impr}[2]{#1 & \textcolor{teal}{\tiny{(+#2)}}}

\resizebox{\textwidth}{!}{%
\begin{tabular}{l *{9}{r@{\,}l}}
\toprule
\multirow{2.5}{*}{\textbf{Model}} & \multicolumn{6}{c}{\textbf{Annual Reports}} & \multicolumn{6}{c}{\textbf{Product Manuals}} & \multicolumn{6}{c}{\textbf{ESG Reports}} \\
[-0.5ex]
\cmidrule(lr){2-7} \cmidrule(lr){8-13} \cmidrule(lr){14-19}
 & \multicolumn{2}{c}{CC} & \multicolumn{2}{c}{OC} & \multicolumn{2}{c}{HC} & \multicolumn{2}{c}{CC} & \multicolumn{2}{c}{OC} & \multicolumn{2}{c}{HC} & \multicolumn{2}{c}{CC} & \multicolumn{2}{c}{OC} & \multicolumn{2}{c}{HC} \\
[-0.5ex]
\midrule

Qwen3.5-397B-A17B & \impr{\textbf{100}}{5.0} & \impr{\textbf{100}}{8.8} & \impr{97.0}{7.9} & \impr{\textbf{100}}{5.9} & \impr{\textbf{98.0}}{2.0} & \impr{97.0}{4.9} & \impr{97.2}{4.0} & \impr{98.5}{4.2} & \impr{\textbf{99.0}}{9.4} \\
GPT-5.5 & \impr{97.2}{10.8} & \impr{98.3}{7.0} & \impr{\textbf{98.5}}{19.1} & \impr{93.0}{12.0} & \impr{94.1}{7.0} & \impr{\textbf{98.0}}{20.8} & \impr{\textbf{99.0}}{13.8} & \impr{\textbf{100}}{11.8} & \impr{96.7}{25.4} \\
Gemini3.1-Pro & \impr{\textbf{100}}{3.3} & \impr{\textbf{100}}{4.5} & \impr{98.2}{2.6} & \impr{97.0}{3.9} & \impr{96.0}{2.9} & \impr{94.1}{3.0} & \impr{98.5}{3.6} & \impr{98.1}{6.8} & \impr{98.1}{4.7} \\
DeepSeekOCR & \impr{96.2}{10.7} & \impr{95.3}{11.0} & \impr{96.1}{13.0} & \impr{94.2}{9.6} & \impr{95.1}{18.7} & \impr{92.1}{12.8} & \impr{96.4}{12.1} & \impr{94.0}{11.8} & \impr{96.1}{17.8} \\
\midrule
Ours (Qwen2.5-VL) & \impr{95.6}{8.7} & \impr{96.9}{15.1} & \impr{96.0}{21.2} & \impr{93.2}{11.8} & \impr{95.5}{17.8} & \impr{93.8}{25.6} & \impr{95.9}{10.0} & \impr{93.5}{12.1} & \impr{95.9}{19.1} \\
\bottomrule
\end{tabular}%
}
\end{table*}

\section{Experimental Setup}

\subsection{Dataset Construction}

We introduce TETHER-Gold, a benchmark comprising 150 recent documents that represent three structural archetypes: (1) Strictly Regulated (50 annual reports, 6,392 pages): Proxying formal, compliance-driven documents like government regulations and industry standards; (2) Stylized \& Unconstrained (50 ESG reports, 3,651 pages): Characterized by high layout entropy, creative typography, and non-linear reading orders, typical of design-centric publications like corporate white papers and visual magazines; and (3) Highly Fragmented (50 product manuals, 11,193 pages): Defined by disjointed text flows frequently interrupted by dense tables, warning blocks, and multimodal diagrams, mirroring complex technical documentation such as medical equipment guides and hardware specifications. Furthermore, to ensure cross-lingual adaptability, the corpus is evenly balanced across English, Simplified, and Traditional Chinese.

Five specialists and a meta-reviewer executed structure-to-content mapping, parsing ToCs into full-depth skeletons to localize each node's semantic onset $p_k$. Ghost node boundaries were inferred via latent semantic transitions and topological constraints.  Independent double-annotation yielded an overall block-level Cohen’s Kappa ($\kappa$) of 0.91 (0.98 for explicit anchors, 0.83 for ghost nodes), with discrepancies adjudicated by the meta-reviewer.

\subsection{Baselines}

TETHER is benchmarked against three system categories: (1) \textit{Deep Hierarchical Recovery Models} (MTD~\cite{hu2022multimodal}, CMM~\cite{wang2023scalable}), predicting hierarchies via supervised visual monotonicity biases; (2) \textit{Dedicated Document Parsers} (MinerU2-VLM~\cite{niu2026mineru2}, Marker~\cite{marker2025}, Docling~\cite{livathinos2025docling}, TextIn~\cite{textin_xparse_docs}), engineered for end-to-end extraction; and (3) \textit{General-Purpose Multimodal Models} (Qwen3.5-397B-A17B, GPT-5.5, Gemini3.1-Pro, DeepSeekOCR), serving as strong zero-shot visual reasoning baselines.

\begin{table*}[t]
    \centering
    \caption{Hierarchy recovery performance (\%). }
    \label{tab:main_results_single_col}
    
    \renewcommand{\arraystretch}{1}
    
    \setlength{\tabcolsep}{6pt}
    
    \resizebox{\textwidth}{!}{%
    \begin{tabular}{ll ccc ccc ccc}
        \toprule 
        
        \multirow{2.5}{*}{\textbf{Category}} & \multirow{2.5}{*}{\textbf{Model}} & \multicolumn{3}{c}{\textbf{ESG Reports}} & \multicolumn{3}{c}{\textbf{Annual Reports}} & \multicolumn{3}{c}{\textbf{Product Manuals}} \\
       
        \cmidrule(lr){3-5} \cmidrule(lr){6-8} \cmidrule(lr){9-11}
         & & OA & TEDS & IA & OA & TEDS & IA & OA & TEDS & IA \\
        
        \midrule
        
        \multirow{2}{*}{\makecell[l]{Deep Hierarchical Recovery}} 
        & MDT & 11.0 & 13.1 & 36.3 & 1.4 & 17.4 & 16.5 & 3.6 & 21.4 & 20.8 \\
        & CMM & 7.8 & 22.8 & 35.2 & 4.2 & 26.1 & 29.0 & 7.3 & 29.9 & 32.4 \\
        
        \midrule
        
        \multirow{4}{*}{\makecell[l]{Dedicated Parsers}} 
        & Marker & 82.6 & - & 37.6 & 52.3 & - & 3.8 & 65.3 & - & 15.5 \\
        & Docling & 83.7 & - & 36.4 & 83.7 & - & 16.4 & 78.2 & - & 14.0 \\
        & MinerU2-VLM & 81.7 & - & 32.9 & 86.9 & - & 6.9 & 80.5 & - & 11.8 \\
        & TextIn & 87.3 & - & 39.7 & 89.6 & - & 35.3 & 87.9 & - & 15.3 \\
        
        \midrule
        
        \multirow{4}{*}{\makecell[l]{General-Purpose LMMs}} 
        & Qwen3.5-397B-A17B & 80.7 & 56.0 & 73.0 & 87.7 & 53.1 & 66.4 & 76.9 & 58.4 & 72.0 \\
        & Gemini3.1-Pro & 75.4 & 48.3 & 58.0 & 75.3 & 45.2 & 53.6 & 67.3 & 43.0 & 51.3 \\
        & GPT-5.5 & 87.6 & 55.1 & 73.7 & 85.4 & 50.7 & 64.3 & 82.4 & 63.1 & 75.3 \\
        & DeepSeekOCR & 79.5 & 39.7 & 54.4 & 56.2 & 52.3 & 61.0 & 66.9 & 42.2 & 55.8 \\
        \midrule
        
        \textbf{Ours} & \textbf{TETHER} & \textbf{96.5} & \textbf{89.1} & \textbf{94.5} & \textbf{98.4} & \textbf{88.0} & \textbf{97.7} & \textbf{98.0} & \textbf{84.7} & \textbf{92.4} \\
        
        \bottomrule
    \end{tabular}%
    }
\end{table*}

\subsection{Metrics}

\subsubsection{Hierarchy Extraction}
 
As TETHER's ToC prior suppresses hallucinations and guarantees high precision, we prioritize recall-oriented metrics for structural completeness:

\textit{\textbf{(1)} Content Completeness (CC)}: The ratio of correctly extracted titles to the ground truth total: $\text{CC} = |T_{\text{correct}}| / |T_{\text{total}}|$. 

\textit{\textbf{(2)} Order Consistency (OC)}: $\text{OC} = |\{(t_i, t_j) \in T_{\text{pairs}} \mid \text{idx}(t_i) < \text{idx}(t_j)\}| / |T_{\text{pairs}}|$, where $T_{\text{pairs}}$ are ground-truth ordered title pairs and $\text{idx}(\cdot)$ is the extracted index. Pairs involving unextracted nodes strictly count as errors.

\textit{\textbf{(3)} Hierarchical Consistency (HC)}: Structural topology fidelity: $\text{HC} = |R_{\text{correct}}| / |R_{\text{total}}|$, representing the validly recovered fraction of ground-truth parent-child edges.

\subsubsection{Hierarchy Recovery} 
We employ a coarse-to-fine protocol:

\textit{\textbf{(1)} Order Accuracy (OA)}: Alignment sequential fidelity: $\text{OA} = |\{(i, j) \in P_{\mathcal{G}} \mid \text{pos}(i) < \text{pos}(j)\}| / |P_{\mathcal{G}}|$, where $P_{\mathcal{G}}$ are ground-truth ordered pairs and $\text{pos}(\cdot)$ is the predicted vertical index.

\textit{\textbf{(2)} Tree Edit Distance Similarity (TEDS)}: 
It quantifies topological transformation costs between predicted ($\mathcal{P}$) and ground-truth ($\mathcal{G}$) trees: $\text{TEDS} = 1 - \text{TED}(\mathcal{G}, \mathcal{P}) / \max(|\mathcal{G}|, |\mathcal{P}|)$.

\textit{\textbf{(3)} Insertion Accuracy (IA)}: Boundary alignment precision via a successor signature $\sigma(h)$ (the immediate non-empty downstream unit). IA measures the exact-match proportion for gold headings $h^G$: $\text{IA} = |\mathcal{H}^G|^{-1} \sum_{h^G \in \mathcal{H}^G} \mathds{1}[\sigma(h^G) \equiv \sigma(m(h^G))]$.

\subsection{Implementation Details}

Eliminating multi-model overhead, TETHER executes both skeleton reasoning and agentic insertion via a single Qwen3.5-27B model to ensure secure local deployment on consumer-grade hardware.

\section{Experimental Results \& Analysis}
We demonstrate TETHER's multi-domain robustness, utilizing ablations and stress tests to validate its architectural irreducibility and operational boundaries, respectively.

\subsection{Hierarchy Extraction}
As shown in Table \ref{tab:toc_extraction_results}, our constrained extraction demonstrates strong generalization across three domains, achieving a ~10.5\% average gain over unconstrained baselines. The framework is particularly effective on the most complex structural metric, HC, delivering a ~13.8\% average gain (up to +25.6\%). Functioning as a performance equalizer, it bridges the gap between smaller models and proprietary giants, elevating the overall mean accuracy to >96\% and restricting the standard deviation to ~1.37\%. Ultimately, despite marginally trailing proprietary models, our 27B backbone enables secure, privacy-preserving local extraction without prohibitive costs.

\begin{table}[t]
    \centering
    \caption{Stage-wise breakdown of ToC-body alignment. $|\mathcal{H}|$: Reference ToC nodes; $|\mathcal{C}|$: Candidate body headings. ESG: ESG reports, AR: annual reports, PM: product manuals.}
    \label{tab:alignment_stats_normal}
    
    \scriptsize 
    \setlength{\tabcolsep}{1.5pt} 
    \renewcommand{\arraystretch}{1.2}
    
    \resizebox{\columnwidth}{!}{%
    \begin{tabular}{l c c c c c c}
        \toprule
        \multirow{2.5}{*}{\textbf{Domains}} & 
        \multirow{2.5}{*}{\makecell{\textbf{$|\mathcal{H}|$} \\ \textbf{H1/H2/H3}}} & 
        \multirow{2.5}{*}{\textbf{$|\mathcal{C}|$}} &
        \multirow{2.5}{*}{\scalebox{0.9}{$\displaystyle \frac{|\mathcal{C}|}{|\mathcal{H}|}$}} &
        \multicolumn{3}{c}{\textbf{Alignment Count}} \\
        \cmidrule(lr){5-7}
        & & & & \textbf{Exact} & \textbf{Fuzzy} & \textbf{Agentic} \\
        \midrule
        
        \textbf{PM} & 
        \makecell{1,465 \\ 355/977/133} & 
        9,335 & 6.37 &
        \makecell{813 \\ (55.5\%)} & 
        \makecell{467 \\ (31.9\%)} & 
        \makecell{185 \\ (12.6\%)} \\
        \addlinespace[2pt]
        
        \textbf{ESG} & 
        \makecell{2,213 \\ 535/1,452/226} & 
        11,028 & 4.98 &
        \makecell{1,438 \\ (65.0\%)} & 
        \makecell{569 \\ (25.7\%)} & 
        \makecell{206 \\ (9.3\%)} \\
        \addlinespace[2pt]
        
        \textbf{AR} & 
        \makecell{1,383 \\ 365/887/131} & 
        3,647 & 2.64 &
        \makecell{1,034 \\ (74.8\%)} & 
        \makecell{246 \\ (17.8\%)} & 
        \makecell{103 \\ (7.4\%)} \\
        
        \bottomrule
    \end{tabular}%
    }
\end{table}

\subsection{Hierarchy Recovery}

We demonstrate TETHER’s SOTA performance, analyze its stage-wise efficacy, and elucidate the mechanisms underlying its noise robustness.

\paragraph{Hierarchy Recovery Performance.}

Table~\ref{tab:main_results_single_col} confirms TETHER achieves SOTA performance (OA $>96\%$, IA $>92\%$) across all domains. This highlights our success in overcoming three systemic bottlenecks in prior art: (1) Deep hierarchical methods rely on rigid heuristics that fail in complex layouts; (2) Dedicated parsers excel at sequential scanning (high OA) but lack hierarchical reasoning (low IA); (3) Standard LMMs suffer from structural drift in long contexts (e.g., annual reports IA drops to 54\%--66\%). TETHER overcomes these limitations by replacing rigid heuristics with four structural constraints, and employing $\Omega$-bounded inference to prevent local errors from cascading into global hallucinations.

\begin{table*}[t]
    \centering
    \caption{Ablation Study of TETHER. Performance and per-document efficiency (total tokens and latency (s)).}
    \label{tab:ablation_study}
    
    \setlength{\tabcolsep}{3.5pt} 
    \renewcommand{\arraystretch}{1.1} 
    
    \resizebox{\textwidth}{!}{%
    \begin{tabular}{l l ccccc ccccc ccccc}
        \toprule
        \multirow{2.5}{*}{\textbf{\#}} &
        \multirow{2.5}{*}{\textbf{Method}} & 
        \multicolumn{5}{c}{\textbf{ESG Reports}} & 
        \multicolumn{5}{c}{\textbf{Annual Reports}} & 
        \multicolumn{5}{c}{\textbf{Product Manuals}} \\
        \cmidrule(lr){3-7} \cmidrule(lr){8-12} \cmidrule(lr){13-17}
        
         & & OA & TEDS & IA & Tok. & Lat. & OA & TEDS & IA & Tok. & Lat. & OA & TEDS & IA & Tok. & Lat.\\
        \midrule
        
        1 & w/o Exact Anchoring & 
        95.5 & 78.9 & 81.3 & 433 & 67.8 & 
        93.1 & 74.1 & 80.9 & 583 & 44.3 & 
        95.9 & 64.2 & 87.3 & 612 & 90.1 \\
        
        2 & w/o Fuzzy Anchoring & 
        83.5 & 63.8 & 53.7 & 266 & 44.1 & 
        86.2 & 63.7 & 71.3 & 577 & 56.0 & 
        80.9 & 62.4 & 42.9 & 536 & 91.7 \\
        
        3 & Fuzzy via String Sim. & 
        92.5 & 67.8 & 83.8 & 205 & 28.7 & 
        88.3 & 63.0 & 87.6 & 538 & 33.4 & 
        91.7 & 61.8 & 85.1 & 420 & 39.7 \\
        
        4 & Fuzzy via Dense Retr. & 
        95.6 & 68.1 & 84.4 & 167 & 24.6 & 
        89.3 & 71.6 & 88.4 & 502 & 38.2 & 
        94.7 & 64.2 & 86.8 & 415 & 41.3 \\
        
        5 & w/o Window ($\Omega_h$) & 
        86.5 & 70.8 & 88.5 & 192 & 35.8 & 
        89.3 & 70.6 & 86.9 & 485 & 46.1 & 
        89.7 & 63.8 & 89.3 & 454 & 51.8 \\
        
        6 & w/o Agentic Insertion & 
        78.3 & 54.5 & 77.9 & - & 0.0 & 
        79.3 & 60.6 & 77.3 & - & 0.0 & 
        64.6 & 63.8 & 62.1 & - & 0.0 \\
        
        7 & w/o Upward Backtrack & 
        83.7 & 71.0 & 83.9 & 172 & 25.4 & 
        86.7 & 61.6 & 85.0 & 353 & 35.1 & 
        84.6 & 64.3 & 80.4 & 307 & 36.5 \\
        
        8 & w/o Top-Down Confirm & 
        95.0 & 74.0 & 74.9 & 138 & 24.3 & 
        91.0 & 71.1 & 81.3 & 198 & 19.9 & 
        93.7 & 72.0 & 83.9 & 157 & 19.7 \\
        
        9 & Direct Context Feed & 
        66.1 & 43.3 & 40.4 & 1140 & 82.8 & 
        47.4 & 47.9 & 40.0 & 2236 & 97.2 & 
        57.5 & 35.2 & 33.5 & 2901 & 89.1 \\
        \midrule
        
        10 & \textbf{TETHER} & 
        \textbf{96.5} & \textbf{89.1} & \textbf{94.5} & \textbf{94.0} & \textbf{16.3} & 
        \textbf{98.4} & \textbf{88.0} & \textbf{97.7} & \textbf{442} & \textbf{32.0} & 
        \textbf{98.0} & \textbf{84.7} & \textbf{92.4} & \textbf{366} & \textbf{30.0} \\
        \bottomrule
    \end{tabular}%
    }
\end{table*}

\paragraph{\textbf{Stage-wise Contribution \& Synergy.}}

Table~\ref{tab:alignment_stats_normal} deconstructs the recall contribution of each processing stage. Symbolic anchoring (exact and fuzzy) establishes a robust baseline, resolving 87.4\% (product manuals) to 92.6\% (annual reports) of unambiguous mappings. Subsequently, the agentic inference layer targets the remaining 7.4\%--12.6\% of intractable cases, such as those involving semantic abstraction or ghost nodes. This synergy ensures that costly LLM reasoning is deployed only when symbolic logic falls short.

\paragraph{\textbf{Mechanism Insight: Robustness to Extreme Noise.}}
A key advantage of TETHER is its resilience in highly fragmented domains (e.g., product manuals), which are characterized by a high $\frac{|\mathcal{C}|}{|\mathcal{H}|}$ ratio (Table~\ref{tab:alignment_stats_normal}) where spurious text artifacts mimic structural headings. While such dense noise induces severe structural drift in baselines (Table~\ref{tab:main_results_single_col}), TETHER exhibits remarkable robustness. Counterintuitively, it repurposes this structural noise by utilizing extraneous candidate headings as frequent, tightly spaced validation anchors. This strategy converts layout density into structural precision, maintaining high alignment accuracy (IA $>92\%$) even within highly irregular product manuals.

\subsection{Ablation Studies}
Table~\ref{tab:ablation_study} validates the architectural irreducibility and computational efficiency of TETHER.

\subsubsection{Performance Analysis}
This section analyzes individual module contributions to alignment accuracy.

\textit{\textbf{Symbolic Anchoring: The Alignment Foundation (rows 1--2).}}
Symbolic matching forms the robust alignment skeleton.
\textit{\textbf{(1)} Exact Anchoring (Row 1):} Its ablation universally degrades structure, confirming its role as an error-cascade stabilizer.
\textit{\textbf{(2)} Fuzzy Anchoring (Row 2):} Its removal causes catastrophic failure, proving it essential for bridging gaps in texts with inconsistent expressions.

\textit{\textbf{Fuzzy Matcher Selection: Why BM25? (Rows 3--4).}}
\textit{\textbf{(1)} String Similarity (Row 3):} Fails due to high-frequency word confusion.
\textit{\textbf{(2)} Dense Retrieval (Row 4):} Yields topological errors via semantic collisions among generic headers.
In contrast, BM25 delivers noise-tolerant lexical granularity to resist semantic collapse, ensuring precise topological gap-filling.

\textit{\textbf{Bounded Inference: Constraints Against Drift (Row 5).}}
Removing the candidate window significantly drops OA, validating that bounded mutual exclusion, enforced by the local candidate window, prevents cross-level mismatches of generic titles, proving essential for topological stability.

\textit{\textbf{Agentic Fine Insertion: Surgical Refinement (Rows 6--8).}}
\textit{\textbf{(1)} Agentic Insertion (Row 6):} Disabling the agent leads to the lowest TEDS scores among pipeline variants, confirming that ghost nodes require semantic inference.
\textit{\textbf{(2)} Upward Backtracking (Row 7):} Its absence results in significant TEDS degradation, as introductory text is misclassified into preceding chapters without this corrective mechanism to rectify boundary shifts.
\textit{\textbf{(3)} Top-Down Confirmation (Row 8):} Critical for precision in data-sparse conditions. Removing this verification step leads to a sharp decline in IA, demonstrating its function in preventing false insertions when local anchor evidence is insufficient.

\textit{\textbf{The Failure of Direct Feeding (Row 9).}}
Directly feeding raw content between anchors to the LLM results in a catastrophic collapse across all metrics. This demonstrates that unconstrained generation within long contexts is severely exacerbated by layout noise and semantic ambiguity, making topological consistency unattainable. In contrast, by refining boundaries via agentic inference with upward backtracking and top-down confirmation, our training-free framework achieves controllable, traceable, and generalizable structure recovery.

\subsubsection{Efficiency Analysis}
Our framework demonstrates superior token and runtime efficiency, as detailed in Table~\ref{tab:ablation_study}.

\paragraph{\textbf{Hierarchical Acceleration.}}
Removing \textit{Symbolic Anchoring} (Rows 1–2) significantly increases runtime and token consumption, highlighting our resource-efficient design: the lightweight symbolic layer quickly resolves high-confidence anchors, thereby bypassing the unnecessary overhead of large models.

\paragraph{\textbf{Impact of Window Constraints on Speed.}}
Applying the constraint window $\Omega_h$ reduces runtime by approximately 50\% compared to Row 5 (w/o Window). This demonstrates that exact matching constraints effectively prune the retrieval search space for fuzzy matching, avoiding unnecessary computational overhead.

\paragraph{\textbf{Targeted Processing vs. Brute-Force Baseline.}}
The Direct Feeding baseline (Row 9) incurs excessive overhead, consuming $>10\times$ tokens (ESG Reports: 1140 vs. 94) and $5\times$ runtime (82.8s vs. 16.3s) versus TETHER. This indicates that unconstrained generation forces the model to process all content between anchors. In contrast, by optimizing boundaries through intelligent reasoning with upward backtracking and top-down confirmation, our framework inputs only essential content: Scenario I evaluates only candidate headings, while Scenario II adds a minimal text stream for backtracking correction.

\subsection{Error Propagation Analysis}
\label{sec:error}

Since our pipeline relies on the document skeleton $\mathcal{H}$ generated during hierarchy extraction, we conducted a stress test to evaluate how these errors propagate to the final hierarchy recovery and prove our metrics ($\text{OA}$, $\text{IA}$, and $\text{TEDS}$) independently isolate distinct structural errors. We injected four types of synthetic noise into $\mathcal{H}$ prior to hierarchy recovery (Table~\ref{tab:error_propagation}):

\begin{table}[t!]
    \centering
    \small
    \caption{Performance degradation (in percentage points, pp) under various types of synthetic noise injected into the intermediate document skeleton $\mathcal{H}$. $\Delta$ denotes the absolute drop in the respective metric.}
    \label{tab:error_propagation}
    
    \renewcommand{\arraystretch}{1.1}
    \setlength{\tabcolsep}{4pt} 
    
    \resizebox{\columnwidth}{!}{%
    \begin{tabular}{l c ccc}
        \toprule[0.4pt]
        \textbf{Noise Type} & \textbf{Ratio} & \textbf{$\Delta$OA} & \textbf{$\Delta$TEDS} & \textbf{$\Delta$IA} \\
        \midrule
        Textual Corruption & 10\% & 2.15 & 17.72 & 1.95 \\
        Node Deletion & 20\% & 2.21 & 11.92 & 2.83 \\
        Depth Randomization & 20\% & 3.44 & 17.72 & 16.72 \\
        Topological Swap & 20\% & 6.30 & 30.97 & 13.45 \\
        
        \bottomrule[0.4pt]
    \end{tabular}%
    }
\end{table}

\textit{\textbf{Textual Corruption}}: Injecting OCR-style errors (e.g., character substitutions, random insertions, or accidental deletions) noticeably impacts $\text{TEDS}$ ($17.72$ pp) but barely affects structural accuracy ($\Delta$OA = 2.15 pp, $\Delta$IA = 1.95 pp). This confirms that superficial mismatches penalize $\text{TEDS}$ solely via its text-substitution mechanics, rather than actual structural degradation.

\textit{\textbf{Node Deletion}}: Simulating extraction omissions yields minor spatial degradation ($\Delta$OA = 2.21 pp, $\Delta$IA = 2.83 pp). Since $\text{OA}$ evaluates relative ordering and $\text{IA}$ anchors on the physical successor $\sigma(h)$, sparse missing nodes fail to disrupt the overarching layout logic, demonstrating considerable tolerance to data loss.

\textit{\textbf{Depth Randomization}}: Perturbing hierarchical levels (e.g., Section to Subsection) severely degrades insertion accuracy ($\Delta$IA = 16.72 pp). Although the relative sequence ($\text{OA}$) remains intact, erroneous depths inherently skew parent-child scoping. This causes the exact-match boundary indicator $\mathds{1}[\sigma(h^G) \equiv \sigma(m(h^G))]$ to fail, highlighting a heavy reliance on accurate depth assignments.

\textit{\textbf{Topological Swap}}: Exchanging structural relationships (e.g., swapping a parent node with its child) inflicts the maximum global degradation ($\Delta$OA = 6.30 pp, $\Delta$TEDS = 30.97 pp). This distortion intrinsically inverts the spatial ordered pairs $\text{pos}(i) < \text{pos}(j)$ essential for $\text{OA}$ and forces expensive global tree-edit operations in $\text{TEDS}$, revealing vulnerability to topological errors, albeit largely mitigated by native priors in practice.

\section{Conclusion}
This study introduces TETHER, which fundamentally shifts the hierarchy restoration paradigm in design-driven long documents away from bottom-up local element aggregation and prediction. By repurposing the ToC into a deterministic structural prior via adaptive anchoring, the framework natively mitigates unconstrained parsing risks and guarantees topological alignment across heterogeneous layouts. Beyond achieving SOTA accuracy and zero-shot universality across three representative domains, extensive evaluations validate TETHER's architectural irreducibility and operational boundaries. Ultimately, with high precision and limited computational overhead, TETHER paves the way for scalable, real-world deployment.

\section*{Limitations}
Fundamentally, this work strictly focuses on design-driven long documents featuring an explicit Table of Contents (ToC). We acknowledge that a small portion of real-world long documents naturally lacks a ToC, rendering TETHER inherently inapplicable to such formats. Furthermore, while our experiments encompass three representative document families—annual reports, ESG reports, and product manuals—across English, Simplified Chinese, and Traditional Chinese, these categories do not exhaust all long-document domains. Finally, although TETHER is a training-free framework, it intrinsically depends on large multimodal and language models for ToC skeleton reasoning and boundary calibration. As a result, its practical application remains bottlenecked by model availability and inference costs and latency constraints.

\section*{Potential Risks}

Although TETHER is designed to recover document hierarchies rather than generate new factual content, its outputs may still pose potential risks. The primary risk lies in error propagation. To quantify this risk, as detailed in Section \ref{sec:error}, we injected four types of synthetic noise into $\mathcal{H}$ prior to hierarchy recovery. These defects can be inherited by the final recovery stage. Consequently, such errors may propagate to downstream applications built upon the structured document, including retrieval, summarization, and decision-support systems. Therefore, when processing topologically degenerate or severely corrupted ToCs, particularly in high-stakes domains such as legal, regulatory, medical, or safety-critical settings, TETHER's structural outputs must not serve as the autonomous basis for decision-making.

\section*{Artifact Use, Licenses, and Data Considerations}

\paragraph{Artifacts and Licenses.}
This work uses several existing scientific artifacts, including document parsers, OCR systems, hierarchical recovery models, and general-purpose multimodal models, as baselines or backbone components. 
We cite the corresponding creators or official references where applicable and use these artifacts only for their intended purposes, namely document conversion, OCR, hierarchy recovery, or multimodal/text inference. 
Open-source tools are used under their respective licenses, while proprietary or hosted models are accessed according to their service terms. 
Our created artifact, TETHER-Gold, is a manually annotated benchmark for evaluating document hierarchy recovery. 
It is intended for research and evaluation purposes only. 
For source documents whose redistribution is not permitted by their original access conditions, we do not redistribute the raw PDFs and instead release only permissible metadata, annotations, or links where allowed.

\paragraph{Intended Use.}
TETHER and TETHER-Gold are designed to support research on structure-aware long-document processing. 
They should be used for evaluating and improving document hierarchy extraction and recovery, especially for ToC-bearing long documents such as corporate annual reports, ESG disclosures, and technical/product manuals. 
When recovered hierarchies are consumed by downstream LLM or RAG applications, users should perform appropriate human verification or domain-specific validation.

\paragraph{PII and Offensive Content.}
The documents used in our benchmark are public-facing corporate reports, ESG disclosures, and technical/product manuals, rather than private user-generated content. 
During annotation and quality checking, annotators inspected the documents for obvious privacy-sensitive or offensive content. 
The annotation target is document hierarchy---section titles, levels, and boundaries---rather than personal attributes or individual-level records. 
Some public corporate reports may mention named executives, board members, or signatories in official roles; such names are part of the publicly released documents and are not used as protected-attribute labels or prediction targets. 
We do not intentionally collect private personal information, and any released annotations focus only on structural information required for hierarchy recovery.

\section*{Acknowledgments}
This study was supported by the National Natural Science Foundation of China (NSFC) under Grant Nos. 72401235 and 72571222, the Natural Science Foundation of Sichuan Province under Grant Nos. 2024NSFSC1061 and 2025ZNSFSC0041, and the Financial Innovation Center of Southwestern University of Finance and Economics under Grant No. FIC2023E007. Additional support was provided by the Artificial Intelligence and Digital Finance Key Laboratory of Sichuan Province, the Ningbo Philosophy and Social Sciences Key Laboratory ``Bulk Commodity Industry Chain and Business Big Data Laboratory,'' and the Zhejiang Provincial Collaborative Innovation Center for Digital Supply Chain and Artificial Intelligence of Bulk Commodities.

\begingroup
\small
\bibliography{sample-base}
\endgroup

\clearpage

\appendix

\setcounter{figure}{0}
\renewcommand{\thefigure}{\thesection.\arabic{figure}}
\renewcommand{\theHfigure}{appendix.\thesection.\arabic{figure}}

\setcounter{table}{0}
\renewcommand{\thetable}{\thesection.\arabic{table}}
\renewcommand{\theHtable}{appendix.\thesection.\arabic{table}}

\section{Challenges in Hierarchical Parsing}
\label{sec:challenges}
Extracting hierarchies from design-driven long documents is hindered by three cascading bottlenecks:

\vspace{4mm}

\begin{figure}[h]
  \centering
  
  \includegraphics[width=\columnwidth]{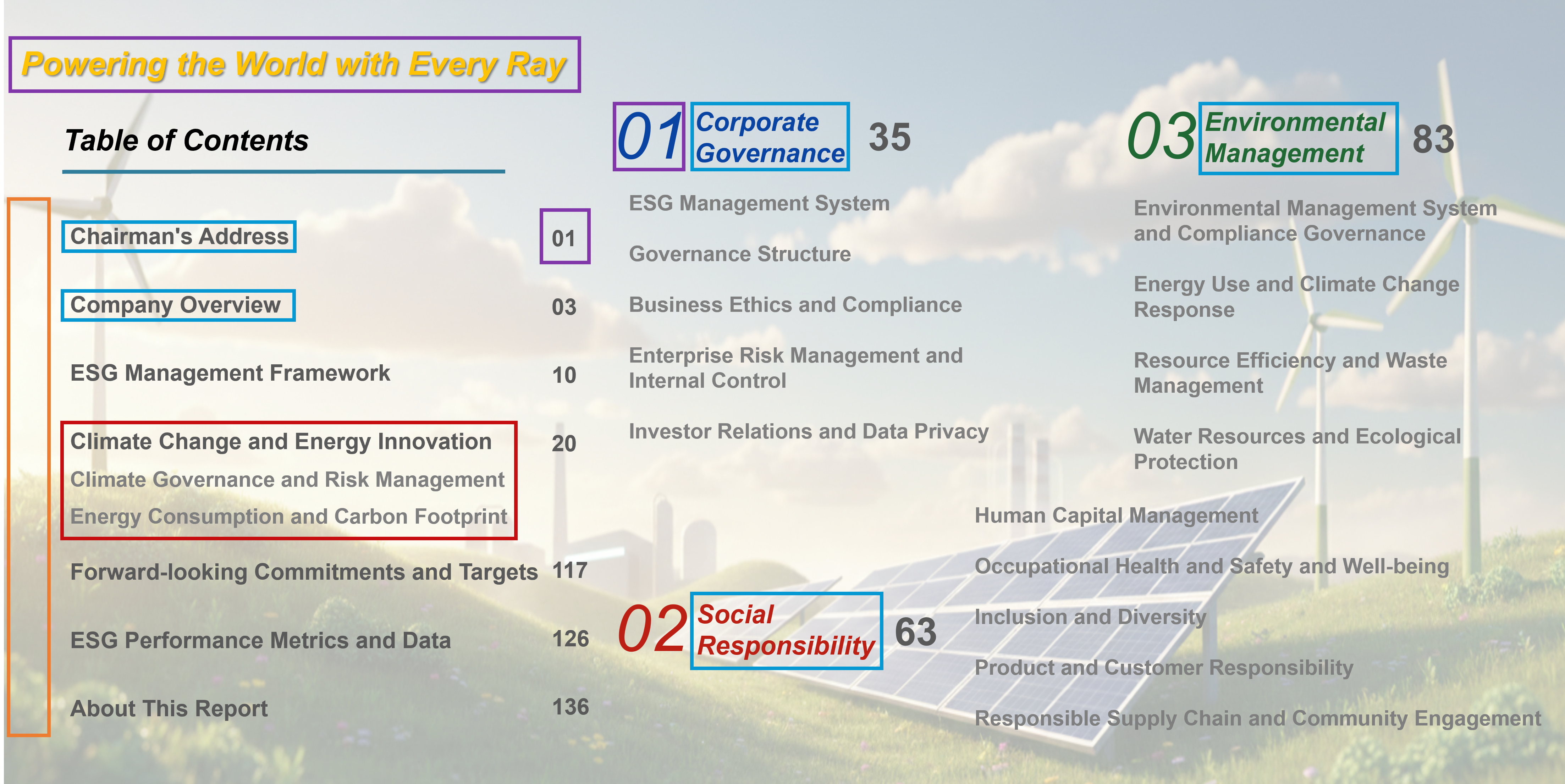}
  \caption{Intrinsic ToC Ambiguity.  (1) Implicit Hierarchy Markers (\textcolor[HTML]{ED7D31}{\rule{2ex}{1.2ex}}). (2) Low Inter-Level Variance (\textcolor[HTML]{C60E11}{\rule{2ex}{1.2ex}}). (3) Stylistic Heterogeneity (\textcolor[HTML]{0598D2}{\rule{2ex}{1.2ex}}). (4) Numerical and Semantic Noise (\textcolor[HTML]{8137AA}{\rule{2ex}{1.2ex}}).}
  \label{fig:toc_challenges}
\end{figure}

\vspace{-4mm}
\paragraph{\textbf{Intrinsic ToC Ambiguity.}} \label{subsec:toc_challenges}
Extreme formatting heterogeneity complicates the reliable extraction of ToC structures (Figure~\ref{fig:toc_challenges}):
\textit{\textbf{(1)} Implicit Hierarchy Markers}: Section numbering is frequently omitted.
\textit{\textbf{(2)} Low Inter-Level Variance}: Visual distinctions between levels are minimal.
\textit{\textbf{(3)} Stylistic Heterogeneity}: Same-level entries exhibit disparate styles.
\textit{\textbf{(4)} Numerical and Semantic Noise}: Numerical polysemy and decorative text induce interference.

\vspace{4mm}

\begin{figure}[h]
  \centering

  \includegraphics[width=\columnwidth]{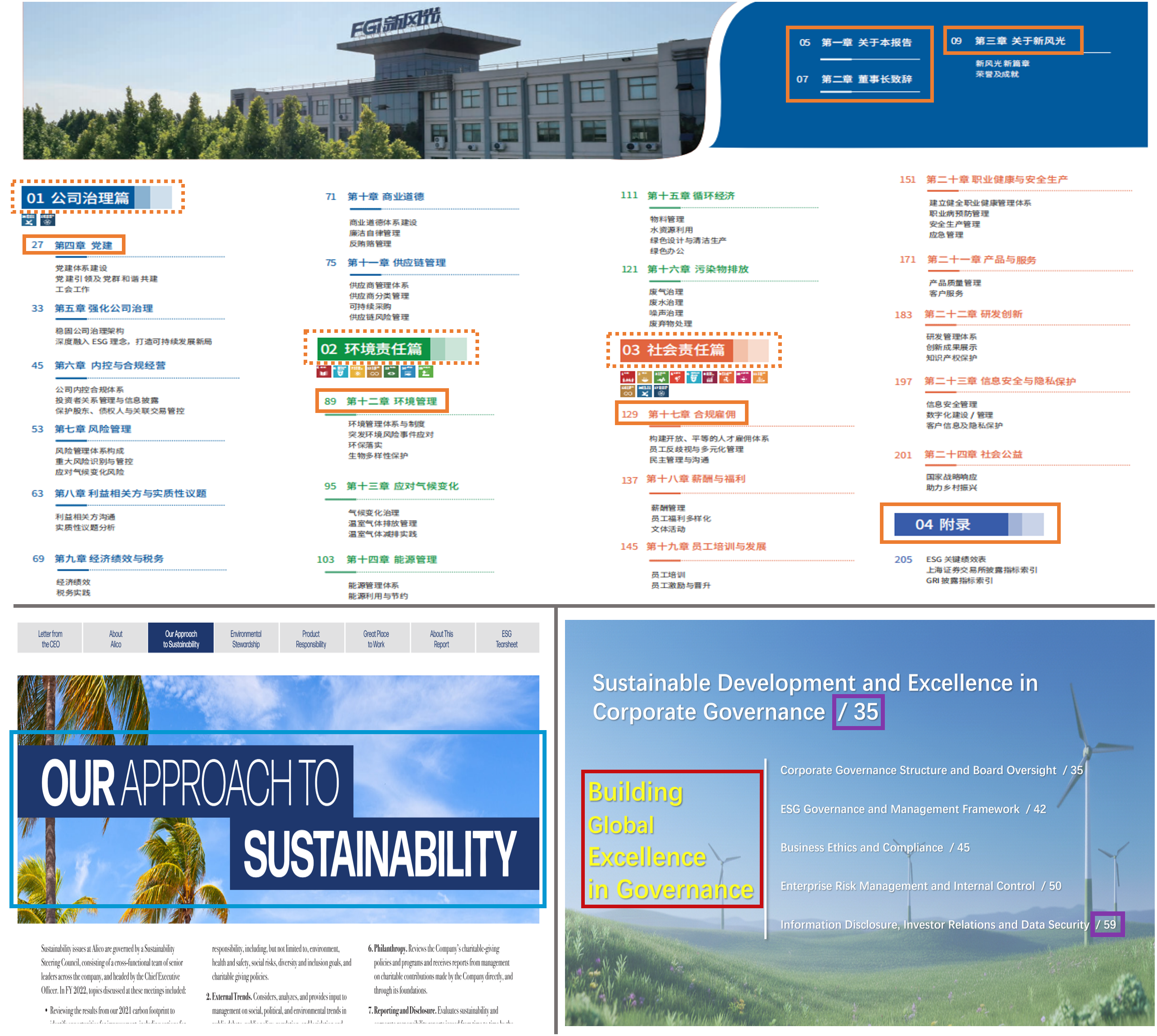}
  \caption{Intra-Body Layout Entropy. (1) Violation of Visual Monotonicity (\textcolor[HTML]{ED7D31}{\rule{2ex}{1.2ex}}), where solid borders denote Level-1 headings and dashed borders mark non-heading content. (2) Visual Decoys (\textcolor[HTML]{C60E11}{\rule{2ex}{1.2ex}}). (3) Text Rasterization (\textcolor[HTML]{0598D2}{\rule{2ex}{1.2ex}}). (4) Structural Discontinuity (\textcolor[HTML]{8137AA}{\rule{2ex}{1.2ex}}).}
  \label{fig:body_challenges}
\end{figure}

\vspace{-4mm}
\paragraph{\textbf{Intra-Body Layout Entropy.}}
Layout entropy decouples visual cues from logical structure (Figure~\ref{fig:body_challenges}): 
\textit{\textbf{(1)} Violation of Visual Monotonicity}: Typographical prominence contradicting logical depth eliminates feature invariance.
\textit{\textbf{(2)} Visual Decoys}: Salient decorative elements mimic logical headers.
\textit{\textbf{(3)} Text Rasterization}: Graphical headers evade OCR, causing node omissions.
\textit{\textbf{(4)} Structural Discontinuity}: Extreme parent-child spans sever lineage dependencies.

\vspace{4mm}

\begin{figure}[h]
  \centering
  \includegraphics[width=\columnwidth]{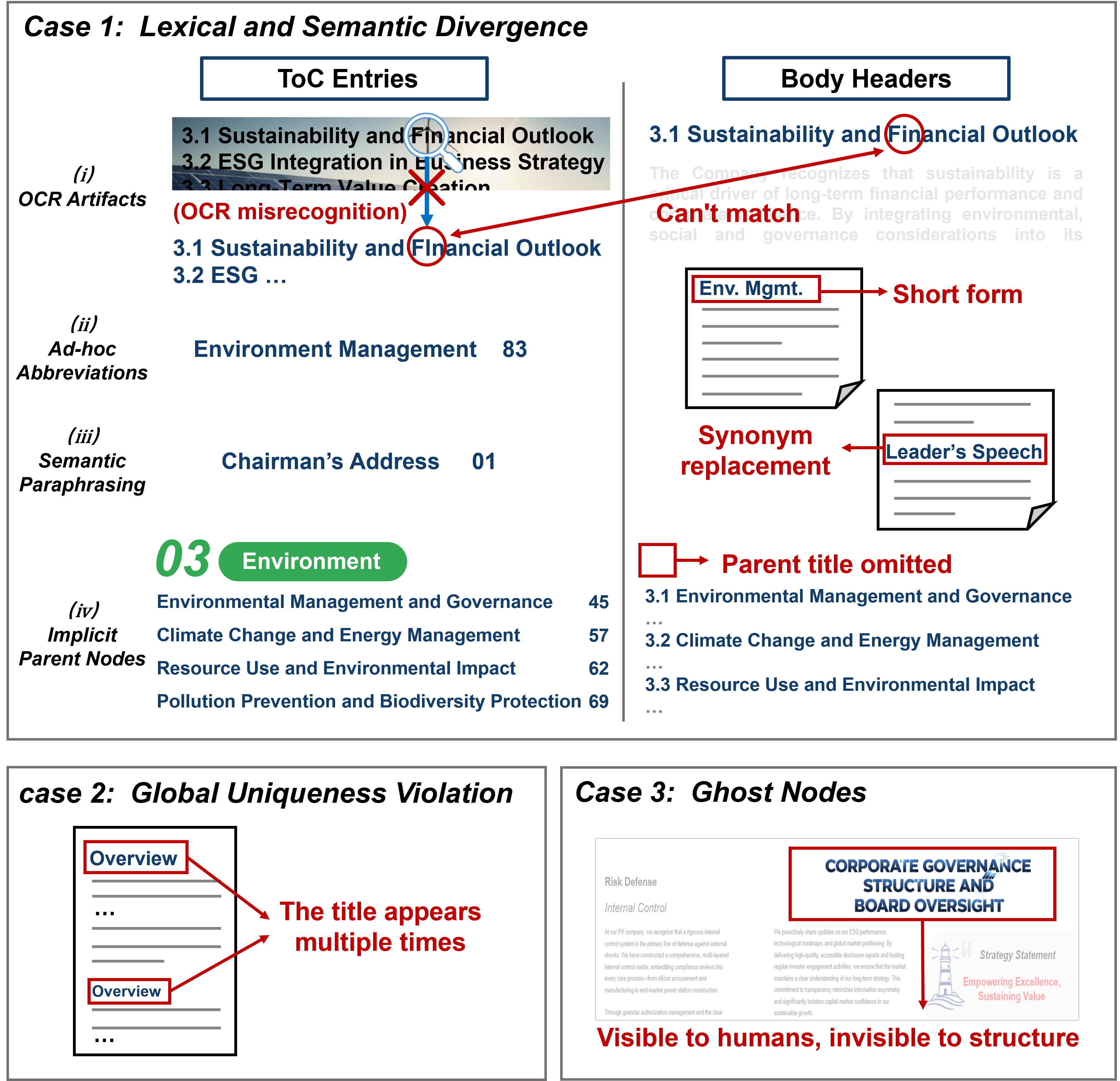}
  \caption{ToC-to-Body Alignment Discrepancies.}
  \label{fig:alignment_challenges}
\end{figure}

\vspace{-4mm}
\paragraph{\textbf{ToC-to-Body Alignment Discrepancies.}}
Both semantic and structural discrepancies confound alignment (Figure~\ref{fig:alignment_challenges}):
\textit{\textbf{(1)} Lexical and Semantic Divergence}: 
\textit{(i) OCR Artifacts}: Character misclassifications defy exact matching.
\textit{(ii) Ad-hoc Abbreviations}: Non-standard contractions expand lexical distance.
\textit{(iii) Semantic Paraphrasing}: Synonymous rewriting yields lexically disjoint headers.
\textit{(iv) Implicit Parent Nodes}: Omitted ToC parents force direct transitions to descendants.
\textit{\textbf{(2)} Global Uniqueness Violation}: Generic headers recur across sections, inducing incorrect anchoring.
\textit{\textbf{(3)} Ghost Nodes}: Rasterized headers lack text, becoming unanchored.

\end{document}